\documentclass[10pt,twocolumn,letterpaper]{article}

\usepackage{iccv}              
\usepackage{afterpage}
\usepackage{bm}
\usepackage{multirow}
\usepackage{float}
\usepackage[accsupp]{axessibility}
\definecolor{iccvblue}{rgb}{0.21,0.49,0.74}
\usepackage[pagebackref,breaklinks,colorlinks,allcolors=iccvblue]{hyperref}

\def\paperID{3967} 
\def\confName{ICCV}
\def\confYear{2025}

\title{CompSlider: Compositional Slider for Disentangled Multiple-Attribute Image Generation}

\author{
Zixin Zhu$^{1,2}$ \quad
Kevin Duarte$^{2}$ \quad
Mamshad Nayeem Rizve$^{2}$ \quad
Chengyuan Xu$^{2}$ \quad
Ratheesh Kalarot$^{2}$ \\
Junsong Yuan$^{1}$ \\
$^{1}$University at Buffalo \quad
$^{2}$Adobe Inc. (ASML) \\
{\small
\texttt{\{zixinzhu,jsyuan\}@buffalo.edu} \quad
\texttt{\{kduarte,mrizve,chengyuanx,kalarot\}@adobe.com}
}
}

\begin{document}
\twocolumn[{
\renewcommand\twocolumn[1][]{#1}
\maketitle
\begin{center}
    \captionsetup{type=figure}
     \includegraphics[width=1.0\textwidth]{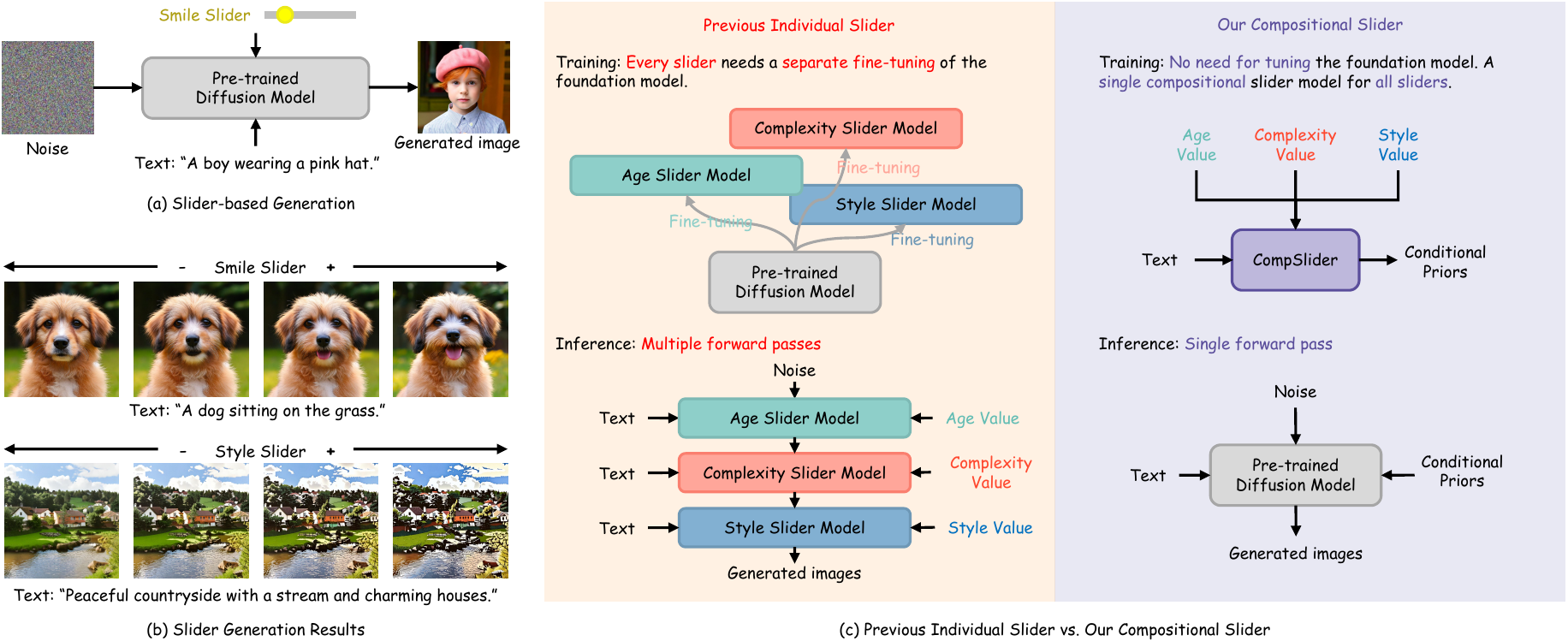}
  \caption{ (a) Illustration of slider-based generation: text defines the primary object, while sliders enable continuous control over specific attributes in the generated image. (b) Examples of fine-grained attribute control in image generation achieved with our slider model. (c) The comparison between previous individual sliders and our compositional slider. Sequential adjustment in previous methods ignores attribute entanglement, leading to unintended interactions. In contrast, CompSlider enables simultaneous control, ensuring better disentanglement and independent adjustments. Pre-trained diffusion model denotes the T2I foundation model.}
  \label{fig:teaser}
\end{center}
}]


\begin{abstract}

In text-to-image (T2I) generation, achieving fine-grained control over attributes - such as age or smile - remains challenging, even with detailed text prompts. Slider-based methods offer a solution for precise control of image attributes.
Existing approaches typically train individual adapter for each attribute separately, overlooking the entanglement among multiple attributes. As a result, interference occurs among different attributes, preventing precise control of multiple attributes together. To address this challenge, we aim to disentangle multiple attributes in slider-based generation to enbale more reliable and independent attribute manipulation. Our approach, CompSlider, can generate a conditional prior for the T2I foundation model to control multiple attributes simultaneously. Furthermore, we introduce novel disentanglement and structure losses to compose multiple attribute changes while maintaining structural consistency within the image. Since CompSlider operates in the latent space of the conditional prior and does not require retraining the foundation model, it reduces the computational burden for both training and inference. We evaluate our approach on a variety of image attributes and highlight its generality by extending to video generation.
\end{abstract}

\section{Introduction}
\label{sec:intro}


\begin{figure}[t]
  \centering
   \includegraphics[width=1\linewidth]{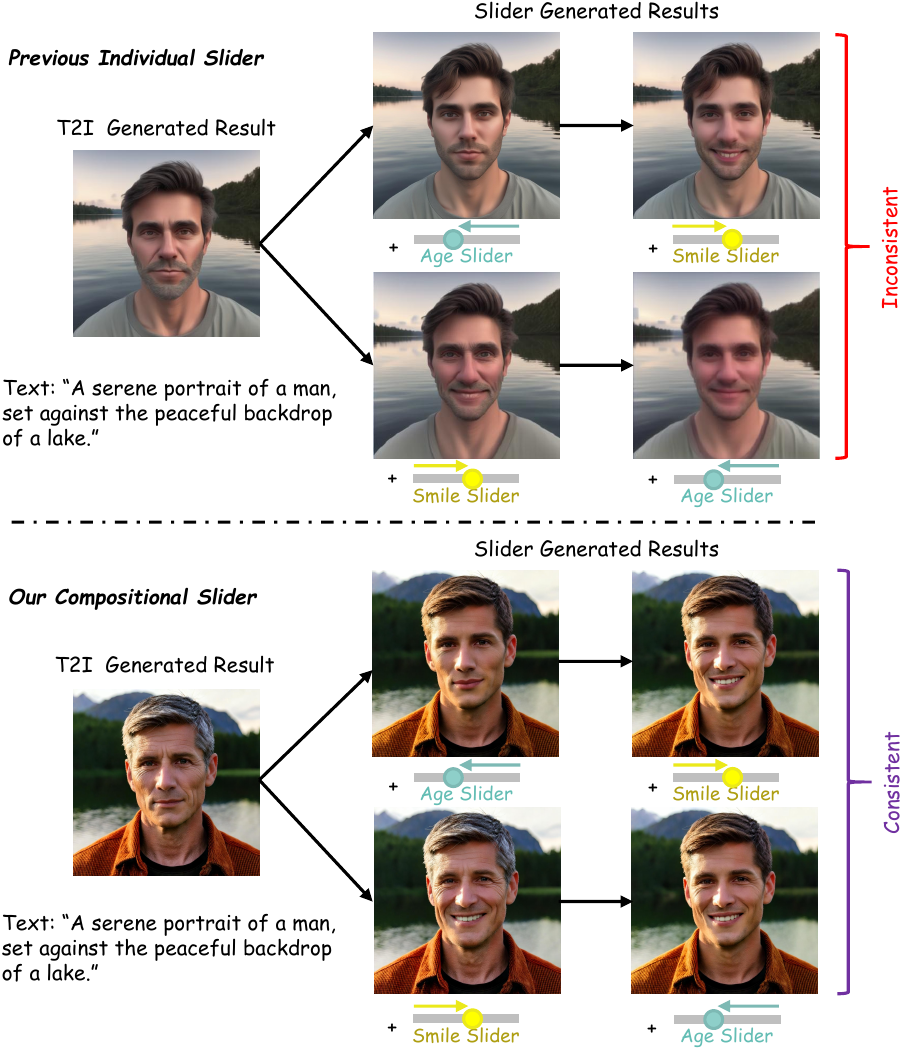}
   \caption{Entanglement in the previous method~\cite{gandikota2023concept} cause a fixed smile value to produce varying smile intensities depending on slider order; applying the smile slider before the age slider yields a young person with a closed smile, while the reverse order results in an open smile. The direction of the arrows indicates the sequential addition of control signals, while text and sliders shown below the images denote the corresponding input signals.}
   \label{fig:onetimemv}
\end{figure}

Text-to-image (T2I) generation~\cite{dhariwal2021diffusion,podell2023sdxl,rombach2022high,ding2022cogview2, feng2025benchmarking, saharia2022photorealistic,ho2020denoising,geng2024instructdiffusion,shi2024dragdiffusion,gu2022vector} made significant progress in recent years. The main avenue of controlling generated outputs involves adjusting input prompts through prompt engineering~\cite{liu2022design,feng2023promptmagician}.  Ideally, a user-friendly approach provides directly and precisely control the attribute generation. 
Although recent methods~\cite{brooks2023instructpix2pix,kawar2023imagic,nichol2021glide,ramesh2022hierarchical,parmar2023zero} to control image generation allow users to specify attributes through text to refine generated images, they lack precise control over attribute strength, e.g. the intensity of an expression, and concepts that are hard to describe in text, such as the complexity of a scene.

Slider-based generation~\cite{gandikota2023concept, sridhar2024prompt} addresses this limitation by extending T2I foundation models: text specifies the main concept, 
while sliders provide auxiliary fine-grained controls, allowing continuous adjustment of various attributes, as shown in Fig.~\ref{fig:teaser}\textcolor{iccvblue}{a}. Such an approach adds flexible, real-time attribute control in this conditional image generation task. Fig.~\ref{fig:teaser}\textcolor{iccvblue}{b} illustrates how sliders enable fine-grained control over a specific attribute in a T2I model. 
Current challenges for slider-based generation include \textit{maintaining continuity}, where slider values accurately reflect the intensity of changes in intended attributes, and \textit{structural consistency} such as identity preservation of a person across attribute adjustments. Previous methods~\cite{gandikota2023concept, sridhar2024prompt} address these challenges by fine-tuning a unique adapter for each attribute and using a guidance scale~\cite{ho2022classifier} to control slider intensity. 


However, images possess several attributes of interest, so users prefer to control multiple characteristics simultaneously. Previous methods overlook the entanglement among multiple attributes, making it difficult to precisely control each attribute while not affect other attributes. As shown in Fig.~\ref{fig:onetimemv}, applying the same smile slider value results in varying smile intensities depending on the age intensities, indicating attribute entanglement. Meanwhile, variations in background and hairstyle further highlight the lack of structural consistency. Moreover, by considering each slider in isolation during training and inference, previous methods face an issue with scalability; for example, if a user wishes to control $N$ attributes in text-to-image generation, then $N$ forward passes would be required, leading to poor user-experience and a large computational burden.

To address these issues and ensure effective disentanglement of inherently independent attributes, we propose CompSlider, a scalable compositional slider model that enables training and inference for multiple sliders by processing all attributes in a single forward pass. CompSlider generates conditional priors~\cite{balaji2022ediff}, \textit{i.e.}, conditioning inputs utilized by a pre-trained T2I foundation model, to produce images that align with user-specified attributes. To train CompSlider, we introduce two novel losses: a disentanglement loss and a structure loss. 
The disentanglement loss enforces that changes in a specific slider attribute do not interfere with other attributes by ensuring the model to generate conditions that accurately represent both commonly occurring and random attribute combinations.
The structure loss regularizes the generated conditional priors within local slider value ranges, preserving structural consistency. 

Notably, CompSlider does not rely on paired data of the same subject with varying attribute intensities (e.g., the same person with both black and red hair), which is costly to obtain. Instead, we leverage random attribute combinations, ensuring robust generalization beyond the biases present in the training data.
Unlike previous methods that require fine-tuning or back-propagating gradients through a T2I foundation model, CompSlider operates within the latent space of the conditional priors, which significantly reduces training costs. Our contributions are summarized below:

\begin{itemize}
    \item We present CompSlider, a conditional image generation approach that allows for fine-grained control of multiple attributes via disentanglement, ensuring independent and precise attribute manipulation. 
    \item We design a training strategy for multiple attribute learning without paired data. Our proposed disentangling and structure losses allow our model to learn arbitrary combinations of attributes that are not commonly found in the training data.
    \item Quantitative and qualitative results show that CompSlider achieves superior continuity and consistency compared to previous state-of-the-art methods. We also show that the approach is generalizable to other modalities, like video.

\end{itemize}

\section{Related Work}


\noindent\textbf{Controlling Image Generation.} Previous controllable image generation~\cite{gafni2022make, nichol2021glide,huang2024pfb,zhu2023designing,zhang2023adding} models have advanced capabilities in personalization, customization, and task-specific image generation. These models offer some degree of control over image properties, with techniques like image diffusion providing adjustments for color variation~\cite{meng2021sdedit,lugmayr2022repaint} and inpainting~\cite{avrahami2022blended,czerkawski2024exploring,huang2024pfb,zhu2023designing,ramesh2022hierarchical}. Approaches like SpaText~\cite{avrahami2023spatext} map segmentation masks to localized token embeddings, enabling more targeted image modifications. Prompt-based image editing~\cite{brooks2023instructpix2pix, huang2023region, tumanyan2023plug} offers practical tools for manipulating images directly through prompts. Different from these methods, we use sliders as auxiliary control signals for text, enabling fine-grained control over various object attributes.
\vspace{-1.5em}
\noindent\paragraph{\textbf{Slider-based Generation.}} Prior controlling image generation methods~\cite{li2023gligen, zhang2023adding, ruiz2023dreambooth, bar2023multidiffusion, bashkirova2023masksketch, mou2024t2i, huang2023composer,gal2022image,li2022diffusion,epstein2023diffusion} often lack fine-grained control over edit strength or complex visual concepts that are difficult to describe in text. To address this, slider-based generation methods have emerged. ConceptSliders~\cite{gandikota2023concept} introduces a low-rank adaptor for enhanced control over concept strength in generated images. PromptSlider~\cite{sridhar2024prompt} uses textual inversion to enable simultaneous control across multiple sliders, but requires each slider to be set to the same value, limiting its practical usability. Unlike these methods, we propose a compositional slider model that enables simultaneous control over multiple attributes, each with distinct values. In addition, in contrast to methods that rely on linear combinations of attribute directions~\cite{li2022stylet2i, preechakul2022diffusion, shi2023exploring}, our approach learns a unified latent mapping that supports compositional control over multiple attributes.

\section{Method}

\begin{figure}[t]
  \centering
  \begin{subfigure}{\linewidth}
    \includegraphics[width=1\linewidth]{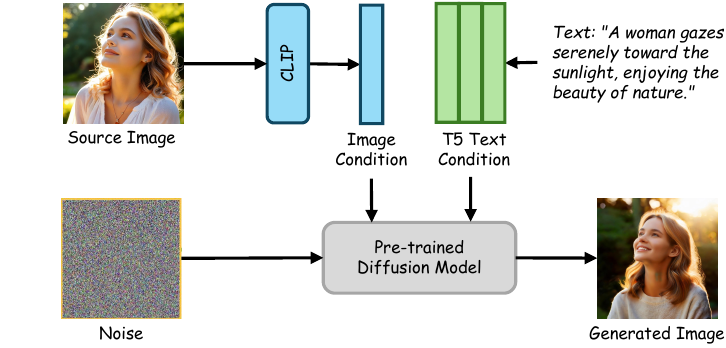}
    \caption{The inference pipeline of our T2I foundation model.}
    \label{fig:method_foundation}
  \end{subfigure}

  \begin{subfigure}{\linewidth}
    \includegraphics[width=1\linewidth]{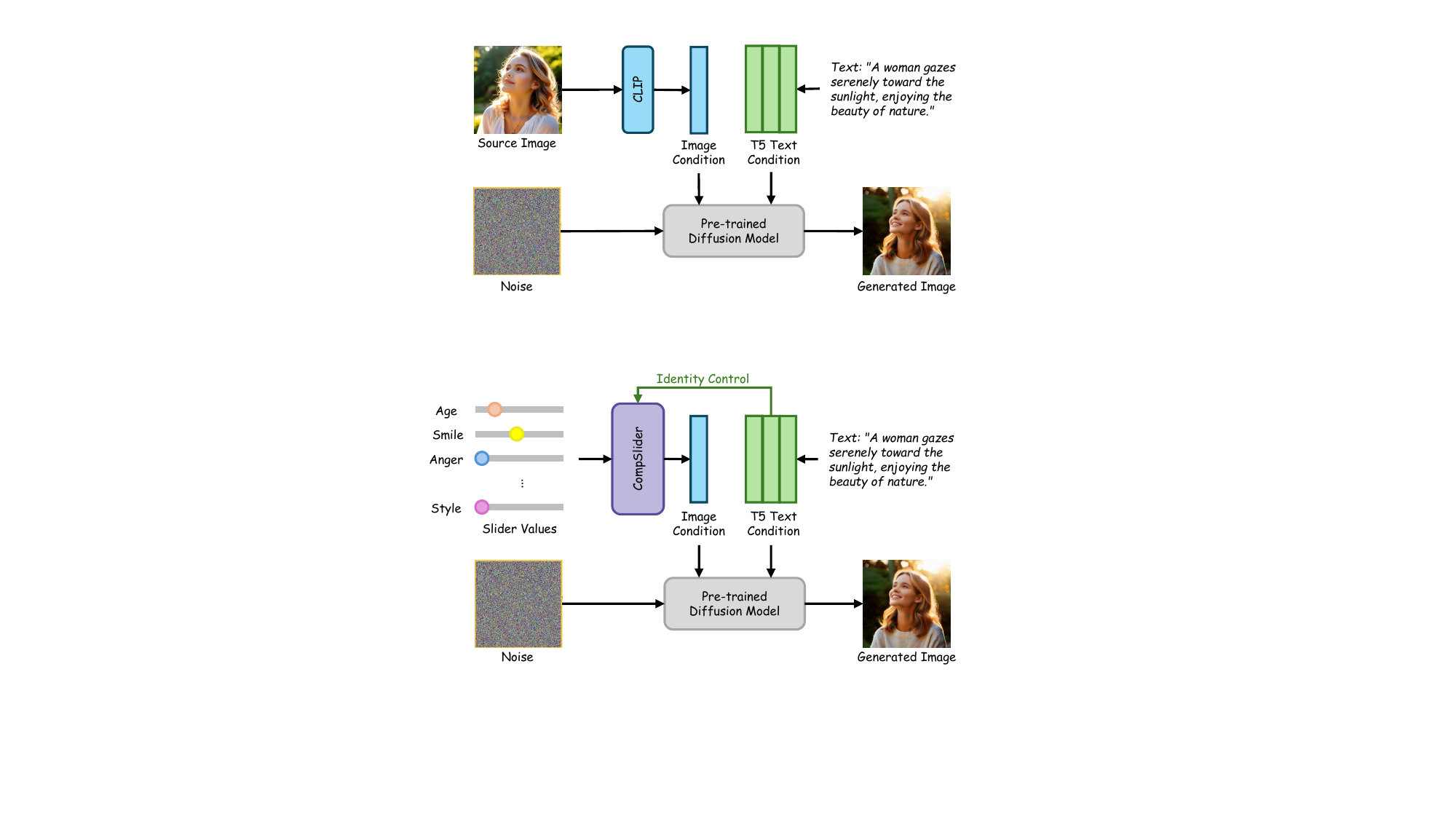}
    \caption{The inference pipeline of our CompSlider.}
    \label{fig:method_ours}
 
  \end{subfigure}

  \caption{In the foundation model, image conditions extracted using CLIP~\cite{radford2021learning} control the style and structure of generated images. In our CompSlider, image conditions are generated based on slider values instead of obtaining from source images.
}
  \label{fig:inference_comparison}

\end{figure}

Given a predefined closed set of $N$ multiple attributes, CompSlider generates multiple-attribute priors as conditions for the pre-trained T2I diffusion model~\cite{rombach2022high,balaji2022ediff} to control various image attributes simultaneously through different sliders. To avoid confusion, we refer to the T2I model as the foundation model in the following sections. Our approach operates without requiring any fine-tuning or gradient backpropagation through the foundation model.
Next, we first elaborate how our CompSlider is integrated with the foundation model and then detail the architecture of it.


\noindent\textbf{T2I Foundation model pipeline.} As shown in Fig.~\ref{fig:method_foundation}, our T2I foundation model is a latent diffusion model similar to Stable Diffusion~\cite{rombach2022high}. But it is trained to support conditioning on both text and optionally on source images for more detailed control like eDiff-I~\cite{balaji2022ediff}. The source image $\bm{x}^{\mathcal{I}}$ and the text $ \bm{x}^{\mathcal{T}}$ are embedded using the CLIP image encoder~\cite{radford2021learning} and the T5 text encoder~\cite{raffel2020exploring} respectively. The image embedding is denoted as $\bm{c}^{\mathcal{I}} \in \mathbb{R}^{dim}$, and the text embedding is denoted as $\bm{c}^{\mathcal{T}} \in \mathbb{R}^{L \times dim}$, where $L$ denotes the length of text token and $dim$ is the dimension of condition vectors. Further details about our foundational model can be found in the Supplementary Materials.


\noindent\textbf{CompSlider pipeline.} 
Motivated by the detailed control~\cite{balaji2022ediff} offered by the input image condition,we design CompSlider to replace the CLIP and original image condition derived from the source image. Compared to the previous foundation model inference pipeline, our approach no longer requires a CLIP encoder or a source image. Specifically, CompSlider takes user-defined slider values and a text prompt as input to generate image conditions $\bm{c}^{\mathcal{I}}$ as multiple-attribute priors. These generated image conditions are then fed into the foundation model to produce final images aligned with the specified slider values. The pipeline of how we apply CompSlider to our foundation model is shown in Fig.~\ref{fig:method_ours}. CompSlider is defined as
\begin{equation}
\bm{c}^{\mathcal{I}} = \text{CompSlider}(\bm{c}^{\mathcal{S}}, \bm{c}^{\mathcal{T}}),
\end{equation}
where $\bm{c}^{\mathcal{S}}$ represents the slider values that control various attributes, $\bm{c}^{\mathcal{T}}$ are the T5 text tokens derived from image captions that define the objects in the generated image.

\subsection{CompSlider}

We employ a Diffusion Transformer (DiT) model~\cite{peebles2023scalable} as our CompSlider, applying the reparameterization~\cite{ho2020denoising} trick to directly predict the image condition. Specifically, instead of predicting the noise $\epsilon$ between \( \bm{c}_t^{\mathcal{I}} \) and \( \bm{c}_{t-1}^{\mathcal{I}} \), the DiT model predicts the pure image condition \( \bm{c}_0^{\mathcal{I}} \). During inference, we use the predicted \( \bm{c}_0^{\mathcal{I}} \) from \( \bm{c}_t^{\mathcal{I}} \) and add noise corresponding to step \( t-1 \) to obtain
\begin{align}
\bm{c}_0^{\mathcal{I}} & = \text{DiT}(\bm{c}^{\mathcal{I}}_t, \bm{c}^{\mathcal{S}}, \bm{c}^{\mathcal{T}},  t), \\
\bm{c}^{\mathcal{I}}_{t-1} & = \sqrt{\alpha_t} \bm{c}_0^{\mathcal{I}} + \sqrt{1 - \alpha_t} \bm{z},
\label{eq:slider}
\end{align}
where  \( \alpha_t \) is the variance schedule term, \(\bm{z} \) is sampled from a normal distribution, $\bm{c}^{\mathcal{I}}_{t-1}$ denotes the denoised image condition at timestep $t-1$, and $\bm{c}^{\mathcal{I}}_t$ is the noisy image condition at timestep $t$. We chose DiT because the image condition itself is a 1024-dimensional vector, which does not require the downsampling operations present in the U-Net. 

\begin{figure}[t]
  \centering
   \includegraphics[width=0.9\linewidth]{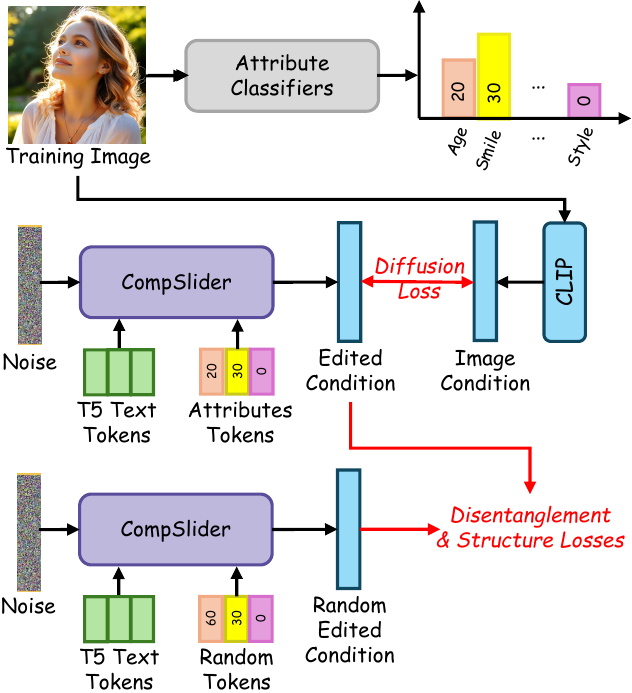}
   \caption{The training process of our CompSlider.}
   \label{fig:priortraining}
 
\end{figure}

\noindent\textbf{Training overview.} The training process for our CompSlider is illustrated in Fig.~\ref{fig:priortraining}. First, slider values are obtained from pre-trained attribute classifiers, embedded as vectors, and fed into CompSlider along with text tokens. The CompSlider outputs image conditions corresponding to these attribute values, ensuring alignment with the latent space of the foundation model..

\noindent\textbf{Embedding slider values.}\label{sec:embeddingslider} For a given pair of text $\bm{x}^{\mathcal{T}}$ and image $\bm{x}^{\mathcal{I}}$, we first use attribute classifiers to obtain classification scores for all attributes, which serve as inputs for slider values. 
We assume that a higher classification score indicates a stronger presence of the attribute in the image, making it suitable as the corresponding slider value. 

Each of the $N$ attribute scores extracted from the image is normalized into the range $[0,1]$ to obtain the slider values, $\bm{v}^{\mathcal{S}} \in [0,1]^N $.
To maintain consistency in the smooth variation of the slider values before and after encoding, we use positional embedding to encode these slider values. The formulation is 
\begin{equation}
\bm{p}^{\mathcal{S}}[j] = 
\begin{cases} 
\cos\left( \bm{v}^{\mathcal{S}} \cdot e^{-\frac{j}{dim/4} \log(f)} \right), & \text{if } j \% 2 = 0 \\
\sin\left( \bm{v}^{\mathcal{S}} \cdot e^{-\frac{j}{dim/4} \log(f)} \right), & \text{if } j \% 2 = 1
\end{cases},
\label{eq:positional_encoding}
\end{equation}
where $\bm{p}^{\mathcal{S}} \in \mathbb{R}^{N \times \frac{dim}{2}}$ is the positional encoding of the input slider values $\bm{v}^{\mathcal{S}}$, $j \in [0, \frac{dim}{2}-1]$ denotes the $j$-th channel, and $f=10000$ controls the frequency of the encoding.

We also introduce a learnable class embedding \( \bm{w} \in \mathbb{R}^{N \times \frac{dim}{2}} \) to ensure the DiT can distinguish between the different attribute sliders. The final slider embedding \( \bm{c}^{\mathcal{S}} \in \mathbb{R}^{N \times dim} \) is obtained by concatenating the positional embedding \( \bm{p}^{\mathcal{S}} \) and the class embedding \( \bm{w} \), defined as
\begin{equation}
\bm{c}^{\mathcal{S}} = [\bm{p}^{\mathcal{S}}, \bm{w}],
\label{eq:class_encoding}
\end{equation}
where \([,]\) denotes concatenation along the channel dimension. Subsequently, the text condition $\bm{c}^{\mathcal{T}}$ and embedded slider values $\bm{c}^{\mathcal{S}}$ are fed into the DiT diffusion model to denoise the image condition $\bm{c}^{\mathcal{I}}_t$.

\subsection{Training Losses}
To train our DiT model, we introduce a novel set of losses: diffusion, disentanglement, and structure losses.

\noindent\textbf{Diffusion loss.}  The diffusion loss ensures the generated conditions match the domain of those from the CLIP image encoder. The formulation is 
\begin{equation}
\mathcal{L}_{\text{diffusion}} = \mathbb{E}_{t, \bm{c}^{\mathcal{I}}_t} \left[ \left\| \bm{c}^{\mathcal{I}}_0 - \text{DiT}(\bm{c}^{\mathcal{I}}_t, \bm{c}^{\mathcal{S}}, \bm{c}^{\mathcal{T}}, t) \right\|^2 \right],
\label{eq:diffusion_loss_x0}
\end{equation}
where $\bm{c}^{\mathcal{I}}_0=\bm{c}^{\mathcal{I}}$ is the original image condition.
However, we observe that using only the diffusion loss can lead to entanglement across different sliders.
As a result, we propose a disentanglement loss and structure loss to disentangle different attributes within our CompSlider. We apply these losses on the reparameterized outputs of the model, predicting \( \bm{c}_0^{\mathcal{I}} \) directly instead of the noise, as the predicted noise itself contains no meaningful information. 

\noindent\textbf{Disentanglement loss.} 
Our proposed disentanglement loss ensures changing individual slider values does not effect other attributes. We posit that the entanglement of multiple attributes originates from the distribution of attribute combinations in the training data, which is inherently imbalanced. Certain attribute pairs appear more frequently than others, leading the model to learn spurious correlations. For instance, people of different ages may exhibit varying degrees of smiling, rather than a uniform distribution across all age-smile combinations. As a result, this bias in the training data leads to entanglement between multiple attributes, where changing one attribute (e.g., age) unintentionally affects another (e.g., smile intensity), making it difficult to achieve precise and independent control. 

To mitigate this issue, we introduce randomly sampled attribute value combinations during training, ensuring that the model does not rely solely on frequently co-occurring attribute pairs. 
In our training process, in addition to the attribute value combinations obtained from training images, we introduce randomly sampling slider values $\bm{v}^{\mathcal{S}*}$ and obtain the corresponding slider embedding, $\bm{c}^{\mathcal{S}*} \in \mathbb{R}^{N \times dim}$, using the method described in Section~\ref{sec:embeddingslider}. 
Since there is no ground-truth image corresponding to the sampled attribute values in this case, we do not apply diffusion loss here. Instead, we ensure that the attribute combinations ($\bm{v}^{\mathcal{S}}$ and $\bm{v}^{\mathcal{S}*}$)
can be accurately recovered from the image conditions generated by CompSlider.
To enforce this, we train an MLP classifier $\mathcal{M}$ which can differentiate between image conditions corresponding to the sampled attribute combinations and the attribute combinations present in the training data.
In this way, we achieve complete multi-attribute disentanglement without relying on paired data with the same subject with varying attribute intensities. 

Specifically, the generated image conditions from the original output embedding, $\bm{c}^\mathcal{S}$, and the ``random" slider embedding, $\bm{c}^{\mathcal{S}*}$ are then concatenated and fed into the the slider classifier to obtain
\begin{equation}
     \{\hat{\bm{s}}_{i}\}_{i=1}^N = \mathcal{M}[ \text{DiT}(\bm{c}^{\mathcal{I}}_t, \bm{c}^{\mathcal{S}}, \bm{c}^{\mathcal{T}}, t),  \text{DiT}(\bm{c}^{\mathcal{I}}_t, \bm{c}^{\mathcal{S}*}, \bm{c}^{\mathcal{T}}, t)],
\end{equation}
such that $\hat{\bm{s}}_{i}\in \mathbb{R}^{B}$ denotes the recovered differences between attribute values obtained from the training images and randomly sampled attribute values, where $i$ denotes the $i$-th slide attribute and $B$ denotes the number of discrete buckets obtained by quantizing the continuous differences. 

For each attribute, the difference between the original and sampled attribute values, \( \bm{v}^{\mathcal{S}} \) and \(\bm{v}^{\mathcal{S}*} \), is calculated as
\begin{equation}
    \{\Delta v_i\}_{i=1}^N = \{v_i^{\mathcal{S}} - v_i^{\mathcal{S}*}\}_{i=1}^N.
\end{equation}
Since we normalize the attribute values to the range [0,1], the differences \( \Delta v_i \) are mapped to the range [-1,1] that is discretized into 
$B$ buckets. Each bucket is treated as a distinct class, to obtain the class label \( y_i \in \{1, ..., B\} \) for the \( i \)-th attribute.
The disentanglement loss \( L_{\text{clss}} \) is defined as
\begin{equation}
    \mathcal{L}_{\text{clss}} = \frac{1}{N} \sum_{i=1}^{N} \mathcal{L}_{\text{CE}} \left( \hat{\bm{s}}_{i}, y_i \right),
\end{equation}
where \(\mathcal{L}_{\text{CE}} \) denotes the cross-entropy loss. 

\noindent\textbf{Structure loss.} To maintain structural consistency across different attribute variations, we introduce a structure loss as a regularization. By ensuring that the predicted image condition remains close to the original, the loss maintains consistency for the same text prompt even when different attributes are applied. To ensure CompSlider can smoothly change attributes within a local range of slider values, we apply the structure loss only when $\Delta v_i$ is below a certain threshold. For our experiments,the threshold is $|\Delta v_i| \leq 0.1$. We formulate the structure loss as
\begin{equation}
\mathcal{L}_{\text{st}} = 
\left\| \text{DiT}(\bm{c}^{\mathcal{I}}_t, \bm{c}^{\mathcal{S}}, \bm{c}^{\mathcal{T}}, t) \right. \left. -  \text{DiT}(\bm{c}^{\mathcal{I}}_t, \bm{c}^{\mathcal{S}*}, \bm{c}^{\mathcal{T}}, t) \right\|_2^2 .
\end{equation}
The final loss \( \mathcal{L} \) is
\begin{equation}
\mathcal{L} = \mathcal{L}_{\text{diff}} +  \mathcal{L}_{\text{st}} + \mathcal{L}_{\text{clss}}.
\end{equation}
Diffusion and structure losses are used to train our DiT model, while disentanglement loss is applied to both the MLP classifier $\mathcal{M}$ and the DiT model.

\begin{table*}[!t]
\centering
\begin{minipage}{0.65\linewidth}
    \centering
    \caption{Quantitative comparison to state-of-the-art slider methods and an editing method Prompt2Prompt~\cite{hertz2022prompt} for human-related sliders.}

    \label{tab:sotahuman}
    \footnotesize
    \begin{tabular}{l|c|c|c|c|c|c}
    \hline
    Method & Cont.\%$\uparrow$ & Cons.\%$\uparrow$ & Scope\%$\uparrow$ & Entang.\%$\downarrow$ & LPIPS$\downarrow$ & CLIP$\uparrow$\\ 
    \hline\hline
    Prompt2Prompt~\cite{hertz2022prompt} & - & 88.47  & 49.46 & 28.99 & 0.19 & 4.15\\
    \hline 
    PromptSlider~\cite{sridhar2024prompt} & 61.17 & 80.23  & 46.25 & 24.31 & \textbf{0.10} & 4.79\\
    ConceptSlider~\cite{gandikota2023concept} & 73.41 & 83.17 & 54.43 & 27.22 & 0.16 & 5.76\\
    CompSlider (ours)                    & \textbf{81.07} & \textbf{90.95} & \textbf{59.02} & \textbf{14.04} & 0.12 &  \textbf{6.20}\\
    \hline
    \end{tabular}
\end{minipage}%
\hfill
\begin{minipage}{0.33\linewidth}
    \centering
    \caption{A/B testing of non-human sliders, comparing the state-of-the-art slider method with our method. The preference probabilities of user are reported.}
   
    \label{tab:sotanonhuman}
    \footnotesize
    \setlength{\tabcolsep}{4.5mm}
    \begin{tabular}{l|c}
    \hline
    Method & Preference\% $\uparrow$ \\
    \hline\hline
    ConceptSlider~\cite{gandikota2023concept} & 34.16  \\
    CompSlider (ours)                     &  \textbf{54.66} \\
    \hline
    \end{tabular}
\end{minipage}

\end{table*}
\section{Experiments}


Since the choice of foundational model depends on specific design choices, each method typically employs a different one. To ensure fair comparison in this section, evaluations are relative rather than absolute. Specifically, metrics focus on comparisons between images generated by the same method under different attribute settings, rather than comparisons across different methods for the same attribute. 

\vspace{-1.4em}
\paragraph{Implementation Details.} 
Our pre-trained attribute classifiers are based on fine-tuned ResNet models~\cite{he2016deep}. For all experiments, our CompSlider model consists of 10 DiT blocks whose input is 128 text condition tokens (from a pre-trained T5-text encoder~\cite{raffel2020exploring}) and 16 slider tokens, corresponding to 16 distinct sliders. All slider types are in the supplementary material. Each tokens are projected to a dimension of $dim=1024$. The total model size is 277M parameters, and it is trained on approximately 3 million images. For our loss calculation, we set the number of buckets $B$ as 20.

\begin{table}[!t]
\centering
\caption{Ablation study on the effectiveness of disentanglement and structure losses.}
\label{tab:abhuman}
\footnotesize
\setlength{\tabcolsep}{1.8mm}
\begin{tabular}{c|c|c|c|c|c|c}
\hline
$\mathcal{L}_{\text{diff}}$ & $\mathcal{L}_{\text{clss}}$ & $\mathcal{L}_{\text{st}}$  & Cont.\%$\uparrow$ & Cons.\%$\uparrow$ & Scope\%$\uparrow$ & Entang.\%$\downarrow$  \\ 
\hline\hline
$\checkmark$ &  &  & 68.96 & 63.21 & 42.06 & 36.68 \\
$\checkmark$  & \checkmark & & 76.49 & 49.29 &  \textbf{63.27} & 19.87 \\
$\checkmark$ & $\checkmark$ & $\checkmark$ & \textbf{81.07} & \textbf{90.95} & 59.02 & \textbf{14.04} \\
\hline
\end{tabular}
\end{table}

\subsection{Human-related Slider Evaluation}


\paragraph{Dataset} We use ChatGPT~\cite{openai2024gpt4o} to generate 300 prompts, each featuring a single person. For each prompt, we generated images with four common emotions, i.e., ``Anger,'' ``Smile,'' ``Surprise,'' and ``Sadness,'' and ``Age,'' resulting in five total sliders. Each slider is sampled at five evenly spaced values, leading to $300 \times 5 \times 5 = 7500$ images. 


\vspace{-1.5em}
\paragraph{Evaluation Metrics} 
Following the common setting~\cite{gandikota2023concept,sridhar2024prompt}, we report the LPIPS and CLIP\footnote{Gandikota \etal \cite{gandikota2023concept} evaluate their model using LPIPS and CLIP scores, but their exact evaluation data are unavailable. Thus, we compute these scores on our evaluation set.} scores between images generated by the foundational model and the slider model to evaluate the quality of generation. But these metrics primarily focused on assessing the semantic correctness of slider-based generation. To consider essential aspects of slider generation, such as maintaining continuity and structural consistency, we propose four new metrics. Our core idea is similar to the re-scoring approach in InterFaceGAN~\cite{shen2020interfacegan}. Specifically, we leverage the widely used open-source human attribute classifier, DeepFace~\cite{serengil2021lightface}, to obtain classification scores for slider attributes in generated images, where a higher score indicates a stronger presence of the attribute. It is worth noting that DeepFace is not the pre-trained attribute classifiers used in our training (Sec.~\ref{sec:embeddingslider}).

These four metrics are continuity, scope, consistency, and entanglement. (1) \textbf{Continuity}: To measure continuity, we check whether images generated with higher slider values receive higher classification scores in DeepFace for the corresponding attribute compared to images with lower values. If this condition is met, it is counted as correct; otherwise, it is counted as an error. The continuity accuracy indicates how precisely each slider controls the target attribute. (2) \textbf{Scope}: The scope score assesses the adjustment range of each slider. We calculate the difference in classification scores between images generated with the maximum and minimum slider values. (3) \textbf{Consistency}: To evaluate structural consistency, we use DeepFace’s recognition model to check if images generated from the same prompt (using different slider values) represent the same individual. Consistency accuracy reflects the model’s ability to preserve identity across different attribute adjustments. (4) \textbf{Entanglement}: This ratio measures the degree of entanglement between different attributes. To compute it, we adjust a single attribute slider while keeping all others fixed. If the classification scores of unaffected attributes change, it is counted as an entanglement sample. A lower entanglement ratio signifies greater independence between attributes, ensuring more precise and isolated control.

\vspace{-1.7em}
\paragraph{Quantitative comparison.} 
Tab.~\ref{tab:sotahuman} contains a comparison with our approach and current state-of-the-art slider approaches~\cite{sridhar2024prompt,gandikota2023concept}. Moreover, although our method falls under the generation task rather than an editing task, we also compare it with Prompt2Prompt~\cite{hertz2022prompt}, as it is an editing method capable of achieving non-continuous attribute control. Benefits from our disentanglement process, as indicated by the entanglement ratio, our approach demonstrates more precise attribute control while better preserving the identity of the generated human faces from the higher continuity and consistency scores, respectively. CompSlider also exhibit the widest adjustment scope, further indicating the model's capability for controlling attributes. 
Moreover, the LPIPS score shows that CompSlider does not negatively impact the foundation model. PromptSlider has better LPIPS , while its significant drop in scope (46.25\% vs. 59.02\%) suggests weaker controllability than CompSlider.


\begin{figure}[t]
  \centering
   \includegraphics[width=1\linewidth]{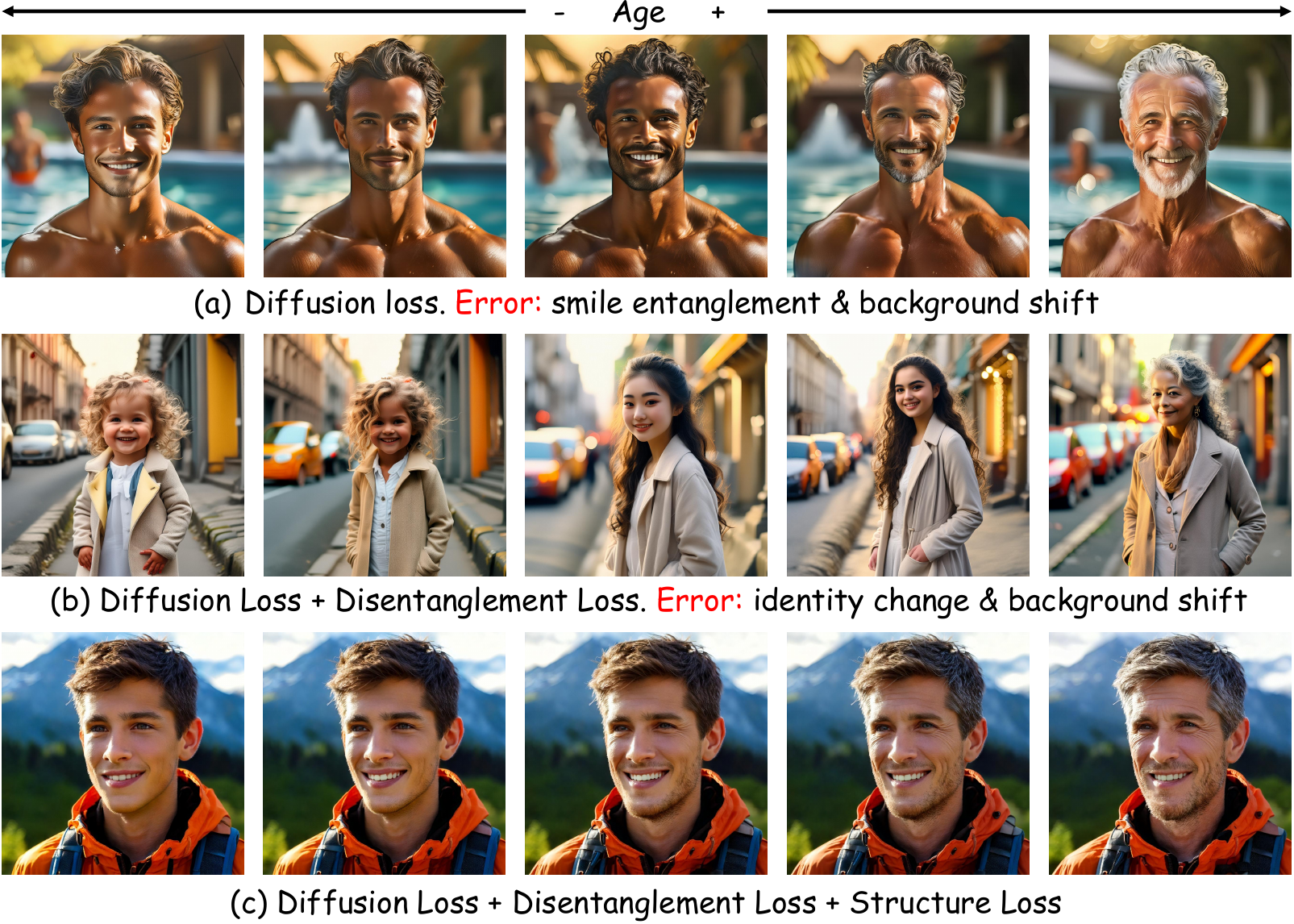}
   \caption{Examples illustrating the effectiveness of the disentanglement loss and structure loss.}
   \label{fig:lossmotivation}
\end{figure}

\begin{figure*}[t]
  \centering
   \includegraphics[width=1\linewidth]{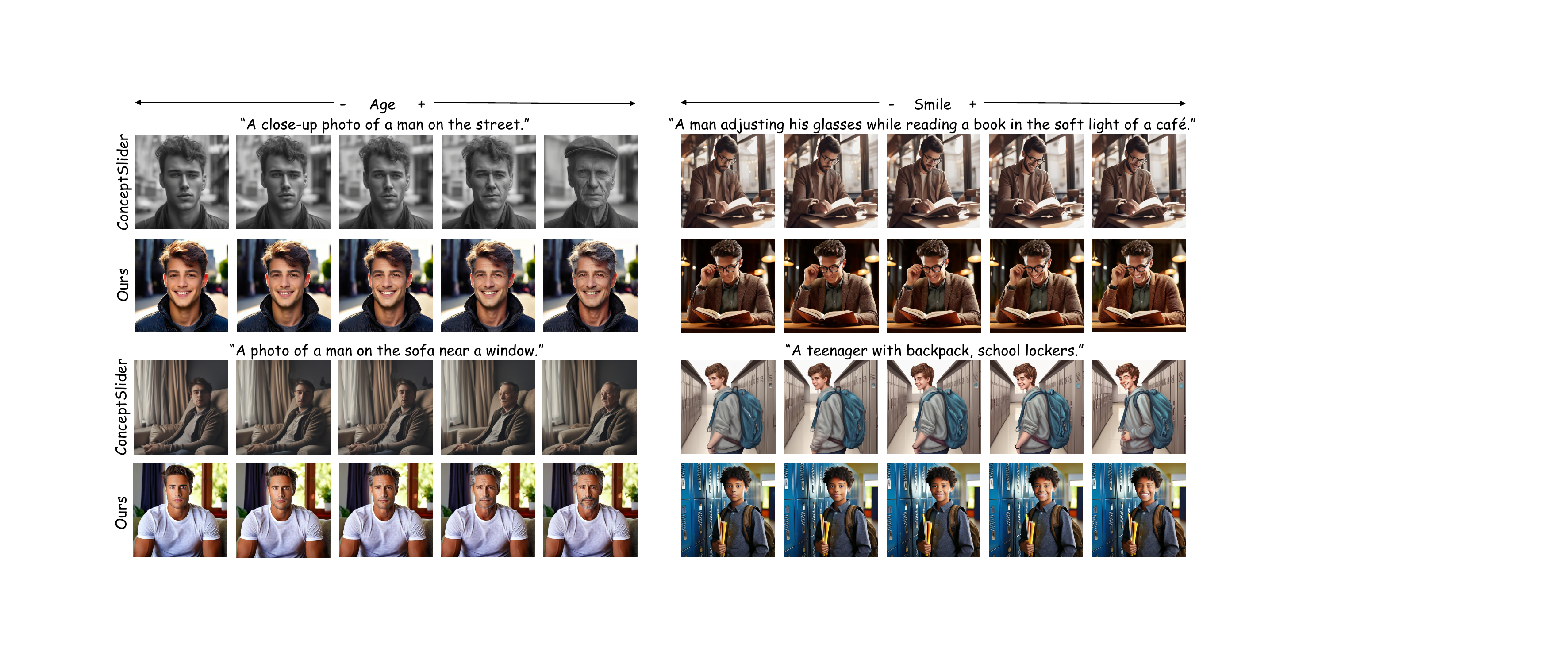}

  \centering
   \includegraphics[width=1\linewidth]{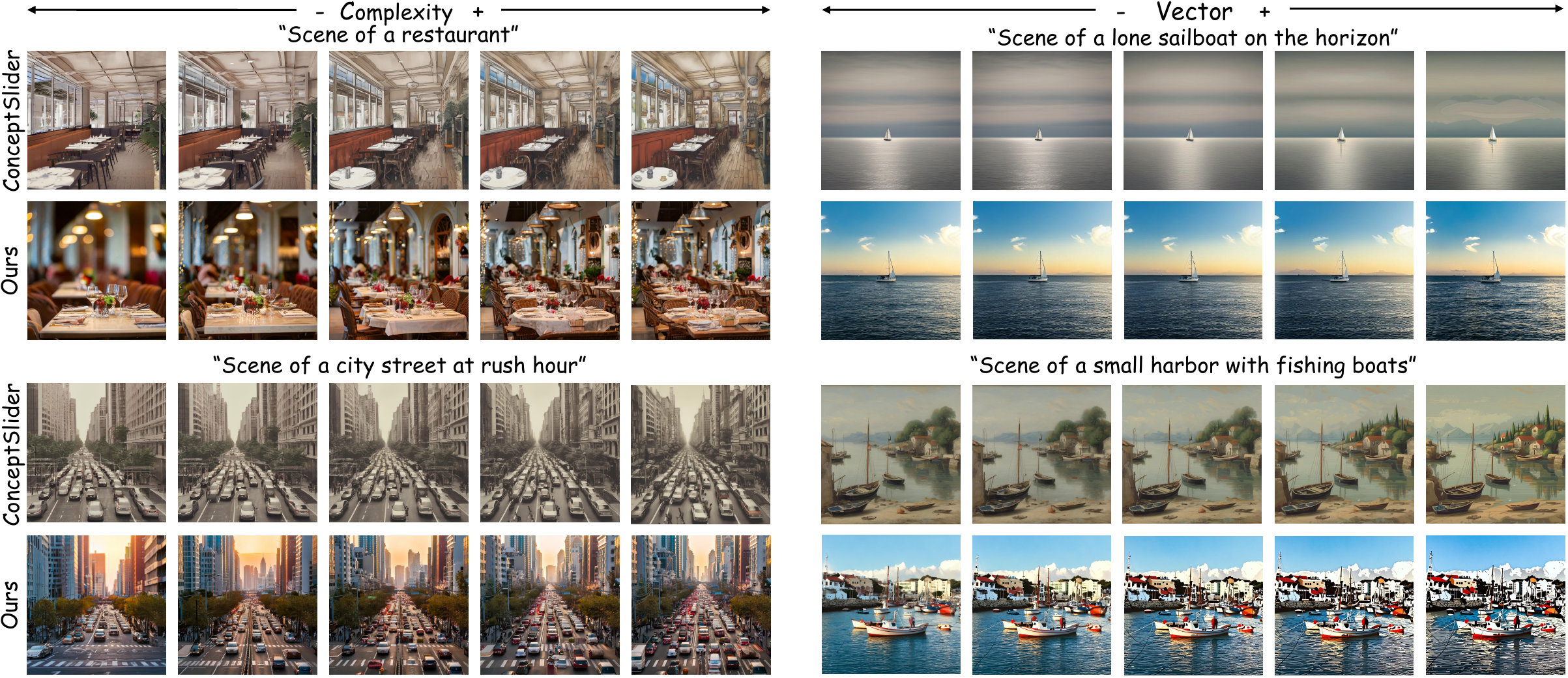}
   \caption{Qualitative Comparison for human-related and non-human sliders. Our results demonstrates better continuity and stronger structure preservation. All compared images are generated from the same text prompt and slider value as those in the upper row. Since slider generation is not an editing task, the initial images generated from the same text prompt are differ.}
   \label{fig:visualnohuman}
\end{figure*}
\begin{figure*}[t]
  \centering
   \includegraphics[width=0.95\linewidth]{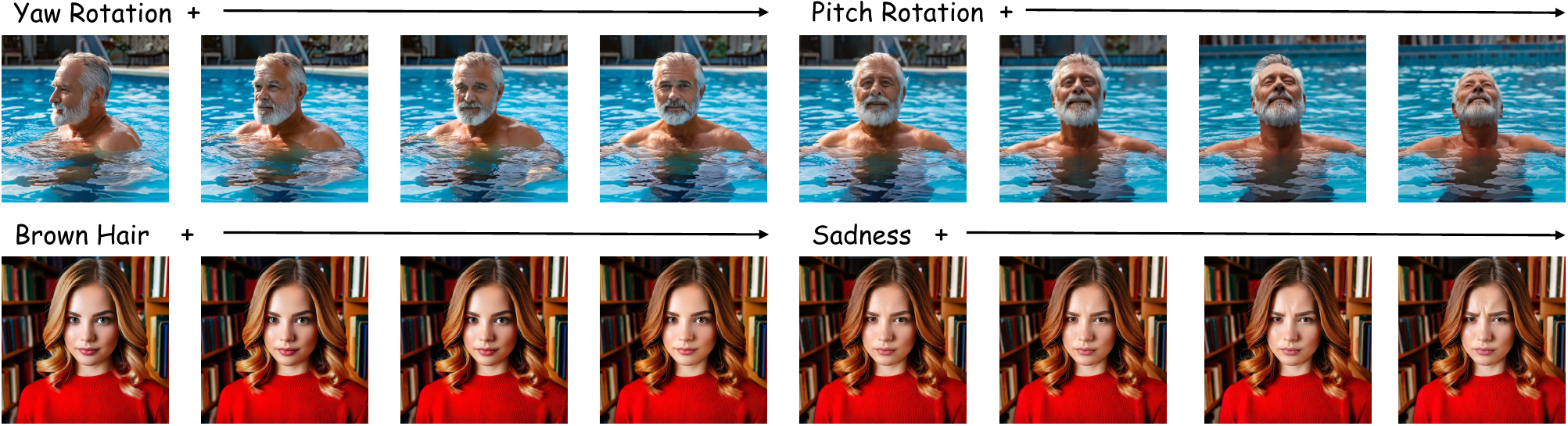}
   \caption{Combinations of different sliders. The Yaw Rotation and Pitch Rotation sliders control the human's horizontal and vertical orientations, respectively, while the Brown Hair slider adjusts the hair color toward brown tones.}
   \label{fig:visualmulti}
\end{figure*}

\vspace{-0.3em}
\paragraph{Successful Disentanglement for Independent Attribute Control.} The reported entanglement ratio in Tab.~\ref{tab:sotahuman} indicates that previous methods suffered from severe multi-attribute entanglement, with the ratio reaching nearly 30\%. After applying our disentanglement approach, this ratio is reduced to 14.04\%, demonstrating the effectiveness of our method in randomly sampling the combination of attributes.

\vspace{-1.7em}
\paragraph{Ablations of Disentanglement and Structure Losses.} To achieve independent attribute manipulation, we introduce disentanglement and structure losses, ensuring that each slider controls its intended attribute without interference. We systematically ablate each loss in Tab.~\ref{tab:abhuman}. With only diffusion loss, the model exhibits entanglement between multiple attributes, resulting in low continuity and scope, along with severe entanglement. After introducing disentanglement loss, $\mathcal{L}_{clss}$, continuity and scope improve significantly, while largely disentangling the attributes, but structural consistency deteriorates. By adding structure loss, $\mathcal{L}_{id}$, the model greatly improves structural consistency (41.66\%), with only a small decrease of 4.25\% in scope. This drop in scope is mainly caused by the regularization effect of the structure loss, which limits the change in the output latent. However, by enforcing identity stability, this regularization further strengthens disentanglement, as a well-preserved identity ensures that attribute variations occur independently without interfering with other factors.  Fig.~\ref{fig:lossmotivation} presents some qualitative examples of the effects of the disentanglement and structure losses.

\vspace{-1.3em}
\paragraph{Qualitative comparison with state-of-the-art method.} 
Fig.~\ref{fig:visualnohuman} presents a comparison between CompSlider and ConceptSlider~\cite{gandikota2023concept} on the two sliders ``age" and ``smile". Both approaches can successfully control the attribute of interest, but our model demonstrates significantly better continuity, control, and consistency: the age and smile smoothly change as their respective slider values are increased, without changing the identity or other aspects of the person.
In contrast, ConceptSlider unintentionally changes the structure. For instance, in the last image for the prompt ``A close-up of a man on the street", ConceptSlider adds a hat to the person when the age slider value is increased. 

\subsection{Non-human Sliders}
\textbf{Datasets and evaluation metrics.} For non-human attributes, since there are no corresponding open-source classifiers, we conduct a human study. We evaluate two different sliders: ``Vector Style" and ``Scene Complexity". Each slider is tested with 100 prompts generated by ChatGPT~\cite{openai2024gpt4o} and five different slider values, resulting in a total of 1,000 images. ``Scene complexity" slider refers to the number of objects present in the scene. ``Vector style" slider refers to transition from the photorealistic style to the vector style. 
Users were asked to take an A/B test to compare our method with ConceptSlider, evaluating the results based on continuity and consistency. For further details about A/B test, please refer to the supplementary materials.

\noindent\textbf{Comparison with state-of-the-art method.} Tab.~\ref{tab:sotanonhuman} shows that CompSlider achieves a user preference probability of up to 54.66\%, while ConceptSlider received only 34.16\% preference. Additionally, users rated 11.18\% of the images as ties, indicating cases where they found both methods equally smooth and consistent. We also present some qualitative results for this comparison in Fig.~\ref{fig:visualnohuman}.

\subsection{Analysis and Applications}

\paragraph{Application to additional attributes.} We also explore various combinations of different attributes. As shown in Fig.~\ref{fig:visualmulti}, CompSlider can control attributes ``yaw rotation", ``pitch rotation", ``brown hair", and ``sadness". Notably our approach works well on attributes which correspond to local changes (e.g. hair color) and global changes. We provide additional visual results in the supplementary material.

\begin{figure}[t]
  \centering
   \includegraphics[width=1\linewidth]{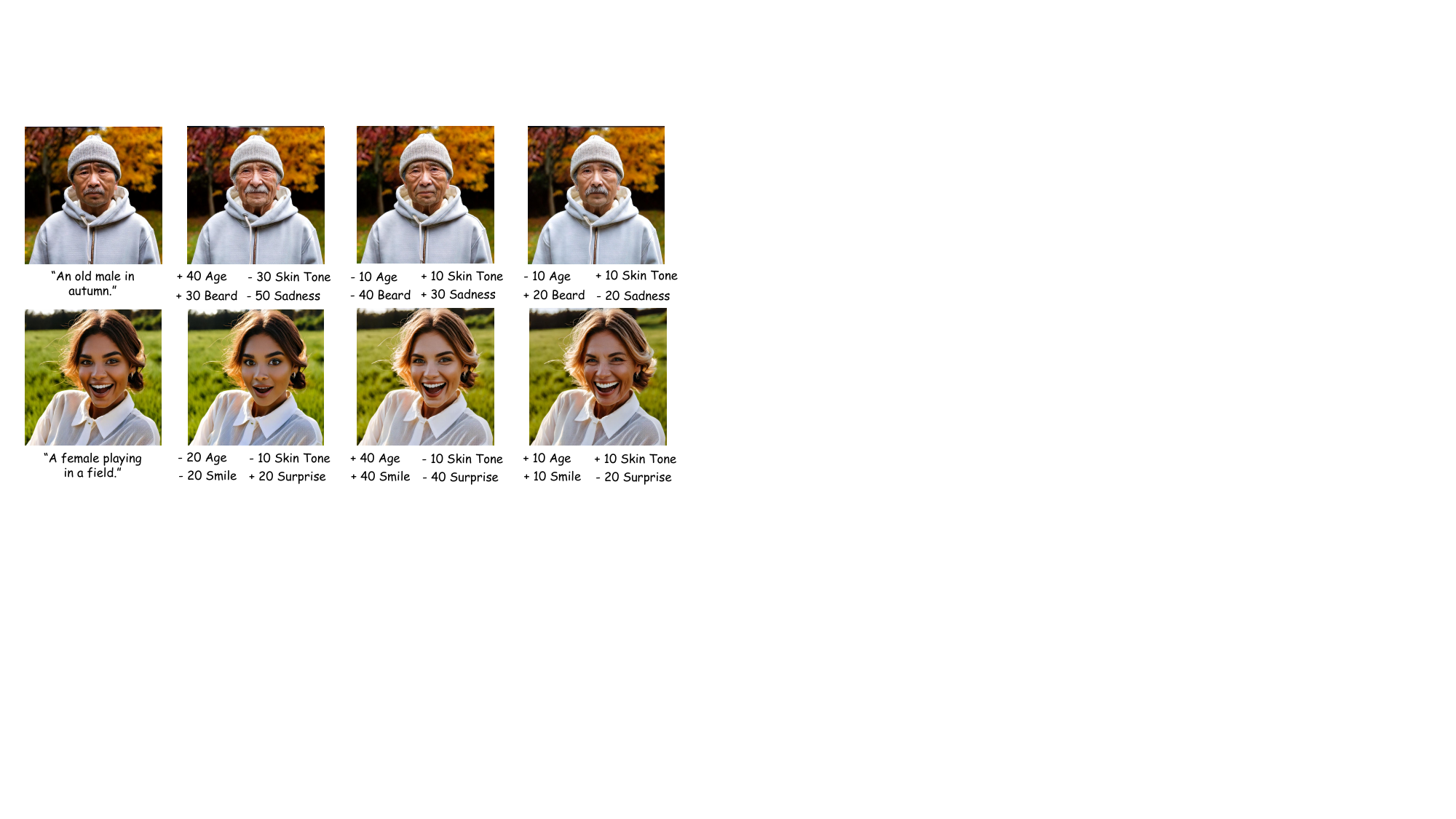}
   \caption{Qualitative results of simultaneous multi-attribute manipulation using our CompSlider.}
   \label{fig:visualonetime}
\end{figure}

\vspace{-1.5em}
\paragraph{Simultaneous multi-attribute control.} Another benefit of our approach is that it supports simultaneous control of all sliders with different values, which is illustrated in Fig.~\ref{fig:visualonetime}. Unlike previous approaches that either allow sequential, single-attribute adjustment~\cite{gandikota2023concept} or limited to changing all sliders by the same amount~\cite{sridhar2024prompt}, our method enables flexible and disentangled adjustments of multiple attributes. 

\vspace{-1.5em}
\paragraph{Extension for Video Generation}

Our method can be easily extended to text-to-video or image-to-video generation, as these video foundation models~\cite{opensora} can utilize CLIP image tokens as conditioning inputs. We demonstrate this extension in the supplementary (Fig.~\ref{fig:visualvideo}), indicating its generalizability.

\section{Conclusion}

This work presents CompSlider, a compositional slider model that generates conditional priors for text-to-image foundation models. By leveraging randomly sampled attribute combinations, our approach ensures effective disentanglement, enabling simultaneous control of multiple attributes while preserving structural consistency. Our quantitative evaluations and qualitative results demonstrate that our model outperforms previous slider-based generation approaches.  A user study validates the controllability of our slider model, confirming its practical effectiveness for diverse attribute manipulation.



{
    \small
    \bibliographystyle{ieeenat_fullname}
    \bibliography{main}
}

\clearpage
\setcounter{page}{1}
\maketitlesupplementary

\noindent In the following section, we first introduce additional details for our foundation model and additional implementation details to provide a deeper understanding of how our slider model operates. We then discuss limitations of our method and explore the impact of threshold selection in structure loss to highlight its role in balancing structure consistency and attribute control. This is followed by a discussion on feature behavior to shed light on how our method effectively guides attribute generation. Finally, we present details of our user studies and showcase additional qualitative results to further validate the effectiveness of our approach.

\section{Additional Details on the T2I Foundation Model}
The foundation model is a U-Net~\cite{ronneberger2015u} that predicts the noise $\epsilon$ between the noisy image \( \bm{x}^{\mathcal{I}}_t \) at timestep \( t \) and the denoised image \( \bm{x}^{\mathcal{I}}_{t-1} \) at timestep \( t-1 \). The single-step inference process is defined as
\begin{equation}
\bm{x}^{\mathcal{I}}_{t-1} =  \frac{1}{\sqrt{\alpha_t}} \left( \bm{x}^{\mathcal{I}}_t - \frac{1 - \alpha_t}{\sqrt{1 - \bar{\alpha}_t}} \text{U-Net}(\bm{x}^{\mathcal{I}}_t, \bm{c}^{\mathcal{I}}, \bm{c}^{\mathcal{T}}, t) \right) \\
 + \sigma_t \bm{z},
  \label{eq:diffusion}
\end{equation}
where \( \alpha_t \) and \( \bar{\alpha}_t \) are the variance schedule terms, \( \text{U-Net}(\bm{x}^{\mathcal{I}}_t, \bm{c}^{\mathcal{I}}, \bm{c}^{\mathcal{T}}, t) \) predicts the noise at time \( t \), \( \sigma_t \) is the standard deviation for the stochastic noise term, and \(\bm{z} \sim \mathcal{N}(0, I) \) is sampled from a normal distribution.  

\section{Additional Implementation Details}
We trained CompSlider using 8 A100 GPUs, and the entire training process took about 16 hours for 20000 iterations. The batch size is set to 2048, and the learning rate is initially warmed up to \(1 \times 10^{-4}\) over 500 steps, then gradually decreased to \(1 \times 10^{-7}\) following a cosine annealing schedule.
Our 16 sliders include: ``Age'', ``Smile'', ``Surprise'', ``Sadness'', ``Anger'', ``Brown Hair'', ``Blond Hair'', ``Gray Hair'', ``Black Hair'', ``Red Hair'', ``Yaw Rotation'', ``Pitch Rotation'', ``Beard'', ``Tone'', ``Vector Style'', and ``Scene Complexity''.
During inference, our CompSlider generates multiple image conditions based on different slider values, which are then input into the foundation model alongside a text prompt to produce the final image results. To ensure consistency across different image conditions for the same text prompt, we keep both the initial noise and the sampled noise in the denoising process of the foundation model identical. Additionally, we found that adding noise (\textit{e.g.}, at an intensity of around 75\%) to the classifier-free guidance~\cite{ho2022classifier} image results, and then using these noise-added images as the initial noise in the foundation model, improves the consistency of details across outputs from different slider values.

\section{Limitations.} When multiple subjects share the same attribute, our method lacks precise target selection, as it applies attribute changes uniformly across all subjects. This limitation arises because the model does not inherently distinguish which subject should be edited. A potential solution to this issue is enhancing the model's text reasoning capabilities, allowing it to better interpret textual instructions and selectively apply modifications to the intended target.

\section{Impact of Threshold Selection in Structure Loss}

\begin{table}[!ht]
\centering
\caption{Ablation Study on the threshold in our structure Loss.}
\label{tab:abinterval}
\footnotesize
\setlength{\tabcolsep}{3.3mm}
\begin{tabular}{c|c|c|c}
\hline
Threshold & Continuity\%$\uparrow$ & Consistency\%$\uparrow$ & Scope\%$\uparrow$ \\ 
\hline\hline
0.5 & 64.68 & \textbf{96.44} & 25.50 \\
0.3   & 77.48 & 92.79  &  57.15 \\
0.1   & \textbf{81.07} & 90.95 & \textbf{59.02} \\
\hline
\end{tabular}
\end{table}

To demonstrate how the threshold in our structure loss affects the slider model's ability to maintain identity, we conducted a search over different threshold values. Tab.~\ref{tab:abinterval} shows that increasing the threshold improves structural consistency. However, a larger threshold comes at the cost of reduced continuity and a narrower adjustment scope, as the structure loss regularization causes the model to produce the same image condition over a wider range of slider values, weakening the sliders' control over attributes.

\begin{figure}[t]
  \centering
   \includegraphics[width=1\linewidth]{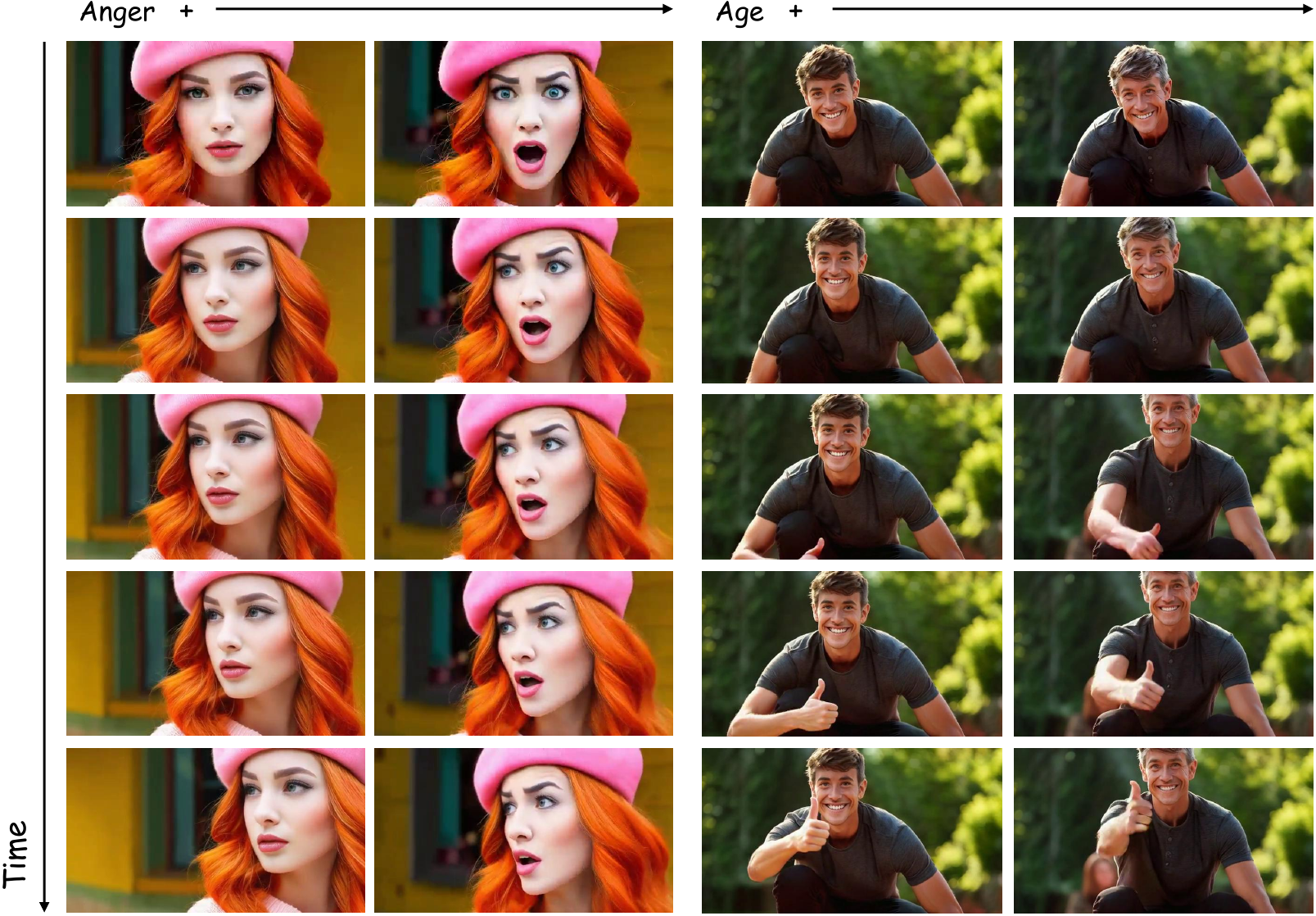}
   \caption{Applying our image slider conditional priors to a video generation model maintains effective control and identity.}
   \label{fig:visualvideo}
\end{figure}

\begin{figure}[t]
  \centering
   \includegraphics[width=1\linewidth]{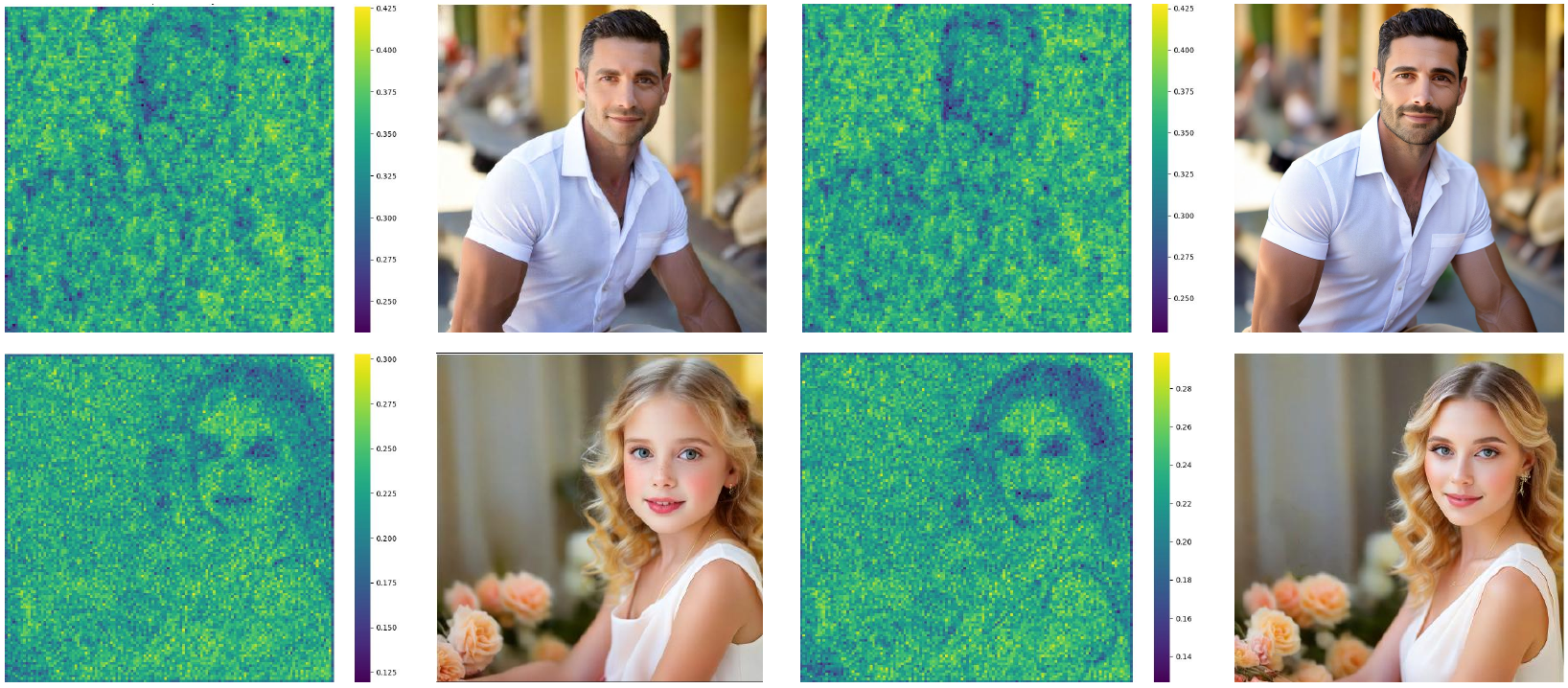}

   \caption{Cross attention in the foundational model between the image conditions generated by CompSlider and the noised image.}
   \label{fig:attnmap}

\end{figure}

\section{Extension to Video Generation}

Figure~\ref{fig:visualvideo} illustrates how CompSlider extends to text-to-video generation using video foundation models. Our method successfully controls the specified attribute consistently across frames, demonstrating its ability to generalize beyond static images.

\section{Discussion on Feature Behavior} 
To further illustrate how our CompSlider generates image conditions and controls the foundational model to produce images with specific attributes, we visualized the cross attention between the generated image conditions and the noised image within the foundational model. Fig.\ref{fig:attnmap} demonstrates how the attention maps from the beard and age sliders guide the model in generating the corresponding output. 

\section{User Studies Details}

\begin{figure*}
  \centering
   \includegraphics[width=1\linewidth]{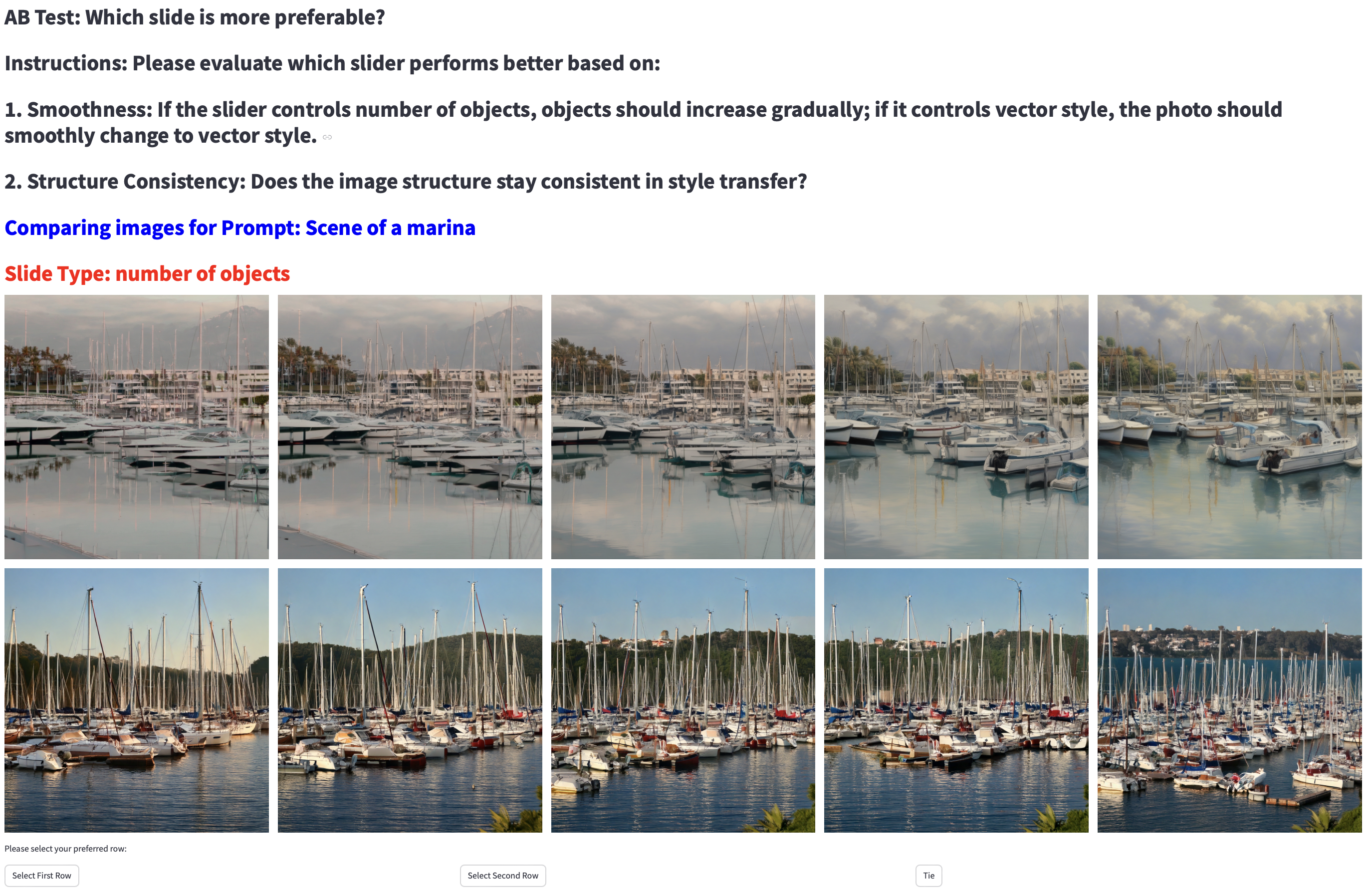}
   \caption{Interface for our A/B test.}
   \label{fig:userstudy}
\end{figure*}

To evaluate our slider method against the current state-of-the-art (SOTA), we conducted an A/B test focusing on smoothness and structural consistency in image transitions. For each test instance, participants viewed a specific slider type (indicated above the images as either ``vector style" or ``number of objects") and two rows of images generated by different methods. The slider intensity increased progressively from left to right, and the order of methods was randomized between the top and bottom rows. Participants selected the row with the smoothest transition and best structural consistency, choosing "First Row," "Second Row," or "Tie." Results were saved to record user preferences across examples. The interface is shown in Fig.~\ref{fig:userstudy}. Similar to common evaluation settings~\cite{hochberg2024towards, liu2024evalcrafter, zhang2024hive}, we conducted a user study with approximately seven participants, all of whom are experts in text-to-image generation research.

\section{Additional Qualitative Results}
\begin{figure*}
  \centering
   \includegraphics[width=0.9\linewidth]{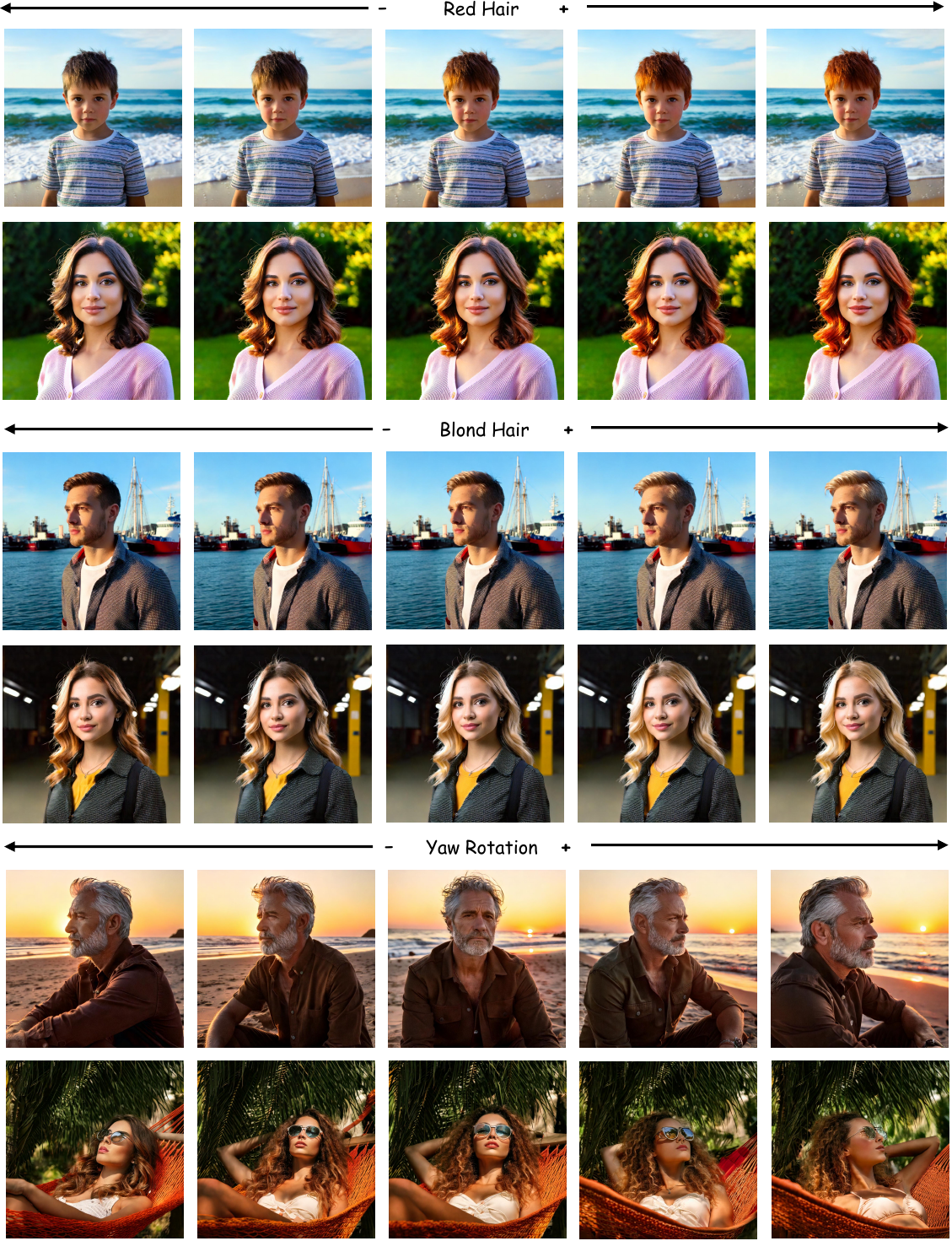}
   \caption{Slider generation results for ``Red Hair", ``Blond Hair", and ``Yaw Rotation".}
   \label{fig:supphuman}
\end{figure*}

\begin{figure*}
  \centering
   \includegraphics[width=0.9\linewidth]{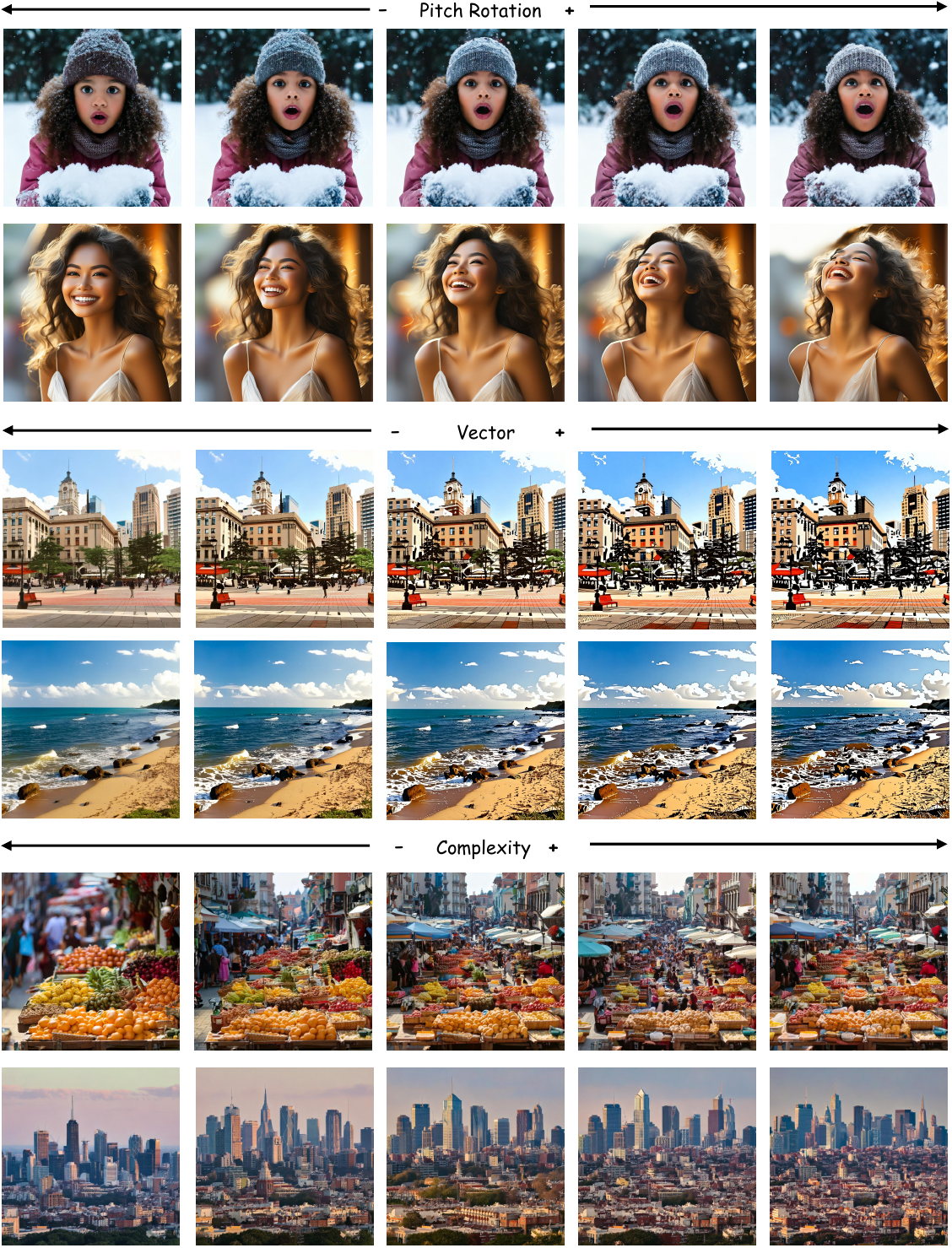}
   \caption{Slider generation results for ``Pitch Rotation", ``Vector Style", and ``Complexity".}
   \label{fig:suppnonhuman}
\end{figure*}

We show additional qualitative examples for sliders in Fig.~\ref{fig:supphuman} and Fig.~\ref{fig:suppnonhuman}.

\end{document}